\documentclass{article}
\usepackage[final]{neurips_2022}
\usepackage{graphicx}
\usepackage{hyperref}
\usepackage{bbding}
\usepackage{siunitx}
\usepackage{booktabs}
\usepackage{subcaption}
\usepackage{enumitem}

\title{Automatic Evaluation of Excavator Operators using Learned Reward Functions}
\author{Pranav Agarwal}
\date{September 2022}

\author{%
  Pranav Agarwal \\
   École de Technologie Supérieure/Mila, Canada\\
  \texttt{pranav.agarwal.1@ens.etsmtl.ca}
   \And
   Marek Teichmann \\
   CM Labs, Canada \\
   \texttt{marek@cm-labs.com}
   \And
   \hspace{55pt}Sheldon Andrews \\
   \hspace{35pt}École de Technologie Supérieure, Canada \\
   \hspace{35pt}\texttt{sheldon.andrews@etsmtl.ca}
   \And
   \hspace{15pt}Samira Ebrahimi Kahou \\
   \hspace{15pt}École de Technologie Supérieure/\\Mila/CIFAR, Canada \\
   \texttt{ samira.ebrahimi-kahou@etsmtl.ca}
}

\begin{document}

\maketitle

\begin{abstract}
Training novice users to operate an excavator for learning different skills requires the presence of expert teachers. Considering the complexity of the problem, it is comparatively expensive to find skilled experts as the process is time-consuming and requires precise focus. Moreover, since humans tend to be biased, the evaluation process is noisy and will lead to high variance in the final score of different operators with similar skills. In this work, we address these issues and propose a novel strategy for the automatic evaluation of excavator operators. We take into account the internal dynamics of the excavator and the safety criterion at every time step to evaluate the performance. To further validate our approach, we use this score prediction model as a source of reward for a reinforcement learning agent to learn the task of maneuvering an excavator in a simulated environment that closely replicates the real-world dynamics.  
For a policy learned using these external reward prediction models, our results demonstrate safer solutions following the required dynamic constraints when compared to policy trained with task-based reward functions only, making it one step closer to real-life adoption. For future research, we release our codebase at \url{https://github.com/pranavAL/InvRL_Auto-Evaluate} and video results \href{https://drive.google.com/file/d/1jR1otOAu8zrY8mkhUOUZW9jkBOAKK71Z/view?usp=share_link}{link}. 
\end{abstract}

\section{Introduction}

Receiving feedback is an important requirement for learning a task. In real life, a student receives this feedback from an expert teacher in the form of a score. While this is a common strategy for teaching a new skill to a novice user, human instructors are often biased~\citep{boysen2009bias}. Furthermore, as the complexity of the task increases, the scores predicted are noisy and have a high variance for users with similar skills. For teaching a novice student a challenging task of excavator automation requires frequent feedback in the form of a score, this is rarely possible with a human expert where the performance is evaluated mostly with a final pass/fail feedback. Similar to a novice student, efficient training of a reinforcement learning (RL) policy,  requires a feedback mechanism such that the agent's actions are evaluated at every time step in the form of a reward from the environment. This form of per-step reward also well known as dense reward requires manually modelling the reward function taking into account the machine dynamics and the safety criteria which is significantly challenging and leads to sub-optimal policies~\citep{zhu2020learning}. Other forms of feedback include a binary sparse reward where similar to human instructor feedback is provided as success or failure at the end of an episode which can lead to slow convergence~\citep{rengarajan2022reinforcement} as the delayed reward leads to a credit-assignment problem. The process of reward formulation thus plays a critical role in the optimization of the policy to reach the best performance~\citep{dayal2022reward}. Hence, automating the reward prediction using inverse reinforcement learning~\citep{Lim2020PredictionOR},\citep{Rosbach2019DrivingWS},\citep{ng2000algorithms} with expert demonstration can overcome the limitations of the above approach. In this work, we explore this direction of learning a score prediction model using demonstration from expert operators, which acts as feedback for teaching a student the task of excavator maneuvering. Since finding new students and setting up the whole system is time-consuming, we validate our feedback model by using it as a reward mechanism for the RL policy (considering the fact that any error in the reward design will lead to incorrect RL policy and unsuccessful task completion) to learn the task of maneuvering an excavator.

\begin{figure}[t]
  \centering
  \includegraphics[scale=0.39]{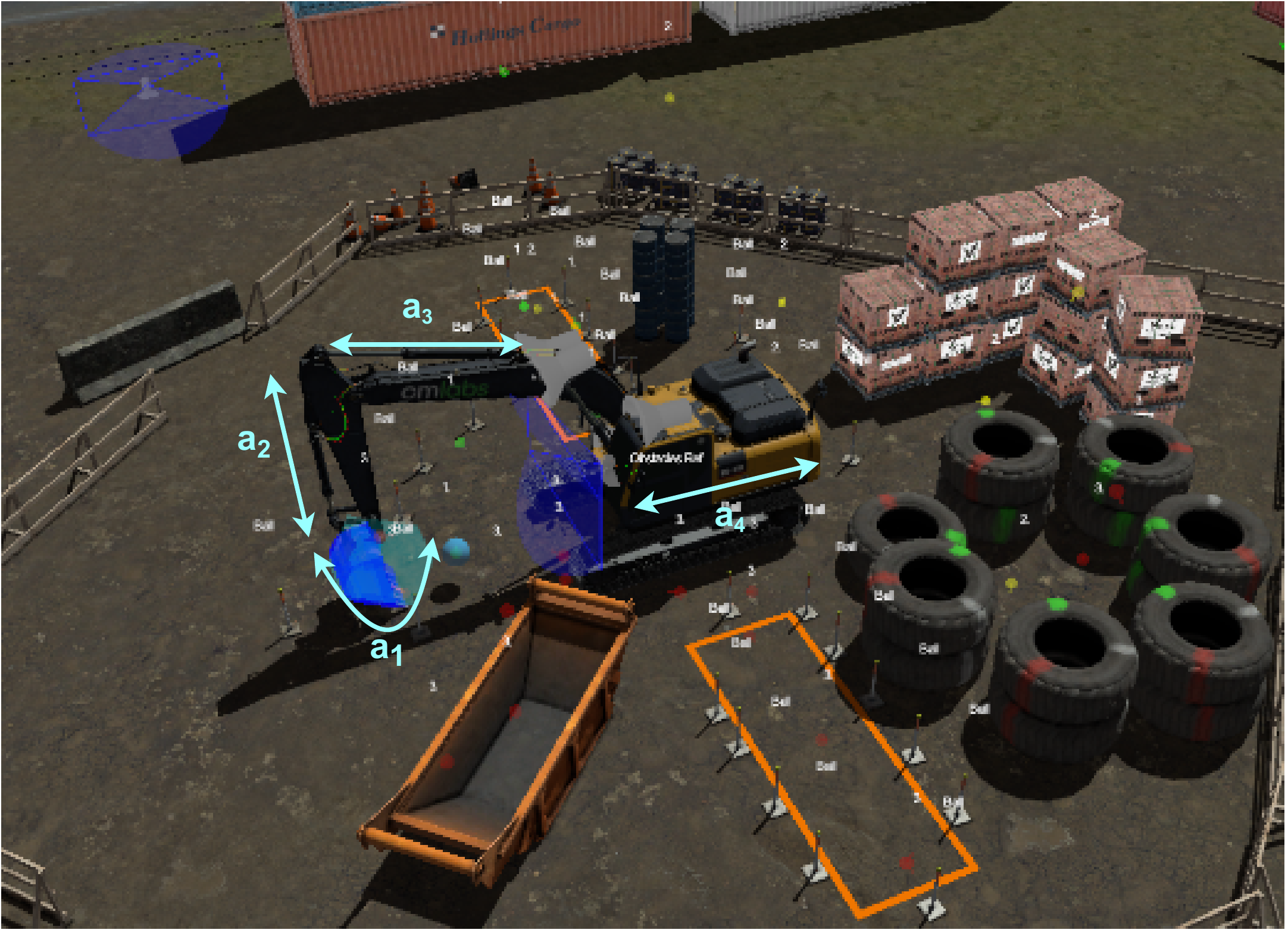}
  \caption{\textbf{Overview of the Vortex environment used for Excavator Maneuvering} The given environment from the Vortex Simulator is used for collecting data from experts, training new users as well as training the RL policy using the proposed automatic score predictive model. In the given environment, the final policy/user needs to reach a fixed goal (based on distance from the start), with precise coordination of multiple actuators that is Bucket ($a_{1}$), Stick ($a_{2}$), Boom ($a_{3}$) and Swing ($a_{4}$), to avoid any infractions while taking into account the dynamic constraints.}
  \label{figurelabel2}
\end{figure}

Excavators have been widely used both in urban as well as rural areas to perform different construction tasks like excavation, loading-unloading and soil levelling. These tasks require careful coordinated control of actuators while taking into account the safety criteria of the surrounding environment. Considering the long working hours with the requirement of precise focus throughout, it impacts the physical and mental well-being of the operators~\citep{la2019we}, impacting the overall performance. Excavator automation is therefore required for rectifying these issues and at the same time improve productivity. Attempts to automate the excavator by manually modelling the controllers \citep{koivumaki2015high}, \citep{zavodni2009actual} for each actuator have been proposed, but are not scalable. Instead, a learning-based approach that considers feedback based on environmental interactions can significantly improve performance. Taking into account this challenging application and in general the sensitivity of an RL policy for a given reward function, we use this application to validate our reward prediction model to guide new trainees.

\begin{figure}[t]
  \centering
  \includegraphics[scale=0.5]{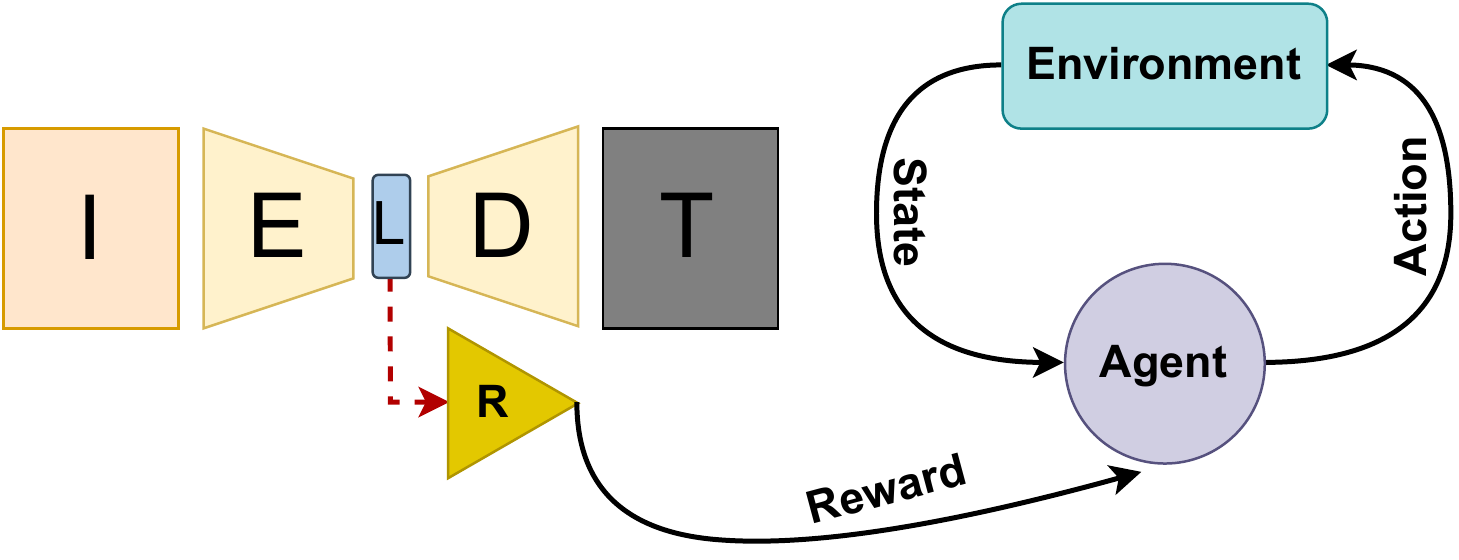}
  \caption{\textbf{Automatic Reward Prediction for Excavator Automation} Overview of the proposed approach for automating the reward prediction for a reinforcement learning algorithm. The approach involves two stages \textbf{a)} Representation learning where Encoder (\textbf{E}) and Decoder (\textbf{D}) learn a distribution (\textbf{L}), for the input (\textbf{I}), depending on the objective (\textbf{T}). \textbf{b)} A reinforcement learning algorithm uses the learned reward function \textbf{R} to solve the task of excavator automation in a simulated environment (Vortex).}
  \label{figurelabel}
\end{figure}

In this work, we learn a score prediction model for evaluating trainees learning to operate heavy construction vehicles, specifically an excavator. We further validate our model by training an RL policy, as shown in Fig.~\ref{figurelabel}, to solve the task of excavator automation. Our proposed reward function is learned following a probabilistic approach using expert data. We show that a distribution taking into account the excavator dynamics and the number of infractions (e.g., unsafe behaviour resulting in collisions with surrounding objects) can act as an ideal source for reward prediction since it efficiently differentiates between actions based on the performance quality. 

Our approach starts by collecting a wide range of diverse user data performing excavator maneuvering in a simulated environment. This dataset consists of excavator sensor readings and per-step infraction values for each user. Taking into account the temporal distribution of these sensor readings, we formulate an objective for learning the dynamics of an expert. Given a sequence of readings as input, the objective is to predict future states using the encoded dynamics. Since an expert and a novice user will have different action dynamics, by learning to predict the next state, the representation learned can efficiently capture these dynamics \citep{epstein2021learning}, further predicting the required reward. To take into account the safety criteria, a different distribution is learned to predict the number of infractions for each time step given the environment information. Thereafter, these reward functions are used to optimize the RL policy for the task of excavator automation in a simulated environment. The main contributions include:
\begin{itemize}[leftmargin=0.5cm]

\item An automated score prediction framework learned using the action dynamics and the number of infraction values of an expert user, which acts as a real-time feedback mechanism for both an external user and the reinforcement learning policy while learning the main task.
\item Demonstration of excavator automation in a simulated environment closely replicating the real world for the task of path maneuvering using the proposed reward function.
\end{itemize}
\section{Related Work}
\textbf{Reward Shaping}. Inspired by human learning, reward shaping is a well-studied problem that involves optimizing an RL policy using an external signal as elaborated in \citep{ng1999policy},\citep{Wiewiora2010}. The current trend mostly involves using a task-based sparse reward \citep{charlesworth2020plangan},\citep{wang2020deep} or manually designing a reward function for optimizing the policy \citep{grzes2017reward}. Sparse reward requires the agent to explore novel strategies \citep{riedmiller2018learning} for solving a task and can lead to slow (or no) convergence \citep{rengarajan2022reinforcement}. A dense (or per step) reward reduces the training time by providing feedback for every action \citep{luo2020balance}, it is usually not straightforward to manually design dense reward functions as the resulting optimal policy is sometimes hard to predict. Considering the simplicity of sparse reward most of the current work uses it over dense reward. \citep{agarwal2021goal} proposes a goal-based sparse reward with a curriculum along with manually designed constraints for the task of autonomous driving. Other strategies include using an intrinsic reward which is a function of state visited as used in \citep{pathak2017curiosity} as additional supervision along with the original reward. These approaches successfully demonstrate novel strategies for different complex tasks. However, none of these works includes subject- and safety constraints based on expert demonstrations of the original reward function leading to unsafe policies. By integrating a score predictive model learned using expert data which is also used as a feedback mechanism for external users, we demonstrate excavator automation for the task of maneuvering in a challenging environment.\\ \\
\textbf{Excavator Automation using Reinforcement Learning}. Excavator Automation is a challenging task that requires precisely coordinated control of multiple actuators in an unstructured environment. Considering the complexity and limited availability of accurate simulators this problem has been rarely explored. \citep{Jud2019AutonomousFT},\citep{Koivo1996ModelingAC} used classical PD and PID controllers for controlling the excavators. Model-based solutions \citep{Dolgov2010PathPF} were proposed which require domain expertise and hence are time-consuming and do not transfer well to new environments. Taking these issues into account the current research is moving towards a learning-based approach. In particular reinforcement learning seems an ideal solution considering its success in solving a wide range of complex robotics manipulation tasks \citep{Chen2021ASF},\citep{Mudigonda2018InvestigatingDR},\citep{Andrychowicz2020LearningDI}. Some of the recent work for excavator automation using RL include \citep{Egli2021AGA},\citep{Egli2020TowardsRH} which learned trajectory tracking for controlling excavator arms by penalizing the policy based on the difference between the expected and the actual trajectory. \citep{andersson2021reinforcementt} used RL for operating forestry cranes for the task of loading-unloading logs using multiple rewards for optimizing grasp success rate, energy, and distance in a curriculum setting. Given the simplicity of the task, each approach relied on manually modelling the reward, limiting its scope to other complex tasks like excavation or maneuvering in a complex environment. In contrast, we focus on optimizing the policy using an expert's demonstration which forces the RL agent to learn the required task while at the same time optimizing the actions to match the expert's dynamic distribution and infractions leading to safer policies following the required constraints.\\ \\
\textbf{Learning Distributions for Reward Prediction.}
To model an expert's action distribution, we learn two independent representations for different features, taking into account the dynamics and the infraction values. This representation is learned following a Variational Autoencoder \citep{kingma2019introduction} like model learning a distribution without any external annotations. Prior works have proposed different strategies for modelling this distribution using auxiliary tasks. \citep{lee2020making} use action-conditioned optical flow prediction of the input RGB images and multiple binary classification losses. This forces the representation to take into account the action dynamics and their impact on the given image which is an important factor while training a reinforcement learning agent. \citep{lesort2018state} describes several representation learning strategies like learning a forward model, using feature adversarial learning or using rewards as objectives. In the current work, we follow a similar strategy as \citep{hu2020probabilistic}, given expert trajectories with multiple sensor states we learn a common representation by predicting future states. This dynamic representation of expert trajectories encodes different excavator states. Using this as a reward function, we optimize the RL policy to complete a goal with a constraint such that the policy's state distribution follows the expert (discussed in detail in Section 4). Further, an independent distribution taking into account the infraction value is also learned to model the safety criteria where the distribution is learned using the objective of predicting the total infraction values given the environment state.

\section{Problem Statement}
This work aims to model a score-predictive model for real-time feedback to excavator operators. To validate this model we use it as a reward function for learning a policy for excavator automation for maneuvering in a complex environment in the presence of multiple external objects as shown in Fig. \ref{figurelabel2}. We study the impact of using different objectives for modelling the distribution of expert trajectories further using it as a feedback mechanism for both an external user and the RL policy. We then evaluate the importance of integrating these rewards as a form of constraint for learning the final policy.

To use the proposed reward model for policy learning, we first learn multiple independent distributions, a probabilistic temporal representation $\textbf{q}_D\textbf{(z)}$ of excavator states given the sampled expert trajectories from the original dataset for modelling the dynamics and $\textbf{q}_I\textbf{(z)}$ of the given environment states at each time step to model infractions. For the dynamic distribution $\textbf{q}_D\textbf{(z)}$ the sequence of \textbf{N} different expert trajectories $\textbf{T}:\{T_0,T_1,...,T_{N-1}\}$ are modelled as a window of $\textbf{x}$ time steps such that for a given time step $\textbf{t}$ the input data is equivalent to $\textbf{I}_t:\{X_t,X_{t+1},...,X_{t+x-1}\}$ and the corresponding label is given as $\textbf{O}_t:\{X_{t+1},...,X_{t+x}\}$. For a given input data $\textbf{I}_t$, our goal is to learn an efficient representation $\textbf{q}_D\textbf{(z)}$ to predict $\textbf{x}$ future states $\textbf{O}_t$ by minimizing the reconstruction loss as shown in Eqn~\ref{eqn1}. The additional KL divergence loss acts as a regularizer such that the learned distribution follows a Normal Distribution. This representation is used to predict the reward by calculating the KL distance between its distribution and the new action (described in detail in Section 4).
\begin{equation} \label{eqn1}
	L_{Dt} = \|X_t - \hat{X_t}\|^{2} + KL(q_D(z_{D})||p_D(z_{D})),
	\end{equation}
with \textbf{$p_D(z_{D})$} representing the prior distribution,
\begin{equation} \label{eqn2}
	p_D(z_{D}) \sim N(0,1)
	\end{equation}
To model the infraction distribution $\textbf{q}_I\textbf{(z)}$ a similar strategy is used with modification in the loss function. At a given time step \textbf{t} the environment states represented as $\textbf{K}_t:\{k^1_t,k^2_t,...,k^f_t\}$ that is the values of \textbf{f} different infractions are used. The distribution is optimized using an objective of predicting the sum of the total infractions $\textbf{S}_t$ and the KL divergence loss as shown in Eqn. \ref{eqn3}. Since the distribution is optimized for the total infractions, the final reward $\textbf{R}_S$ is calculated as the distance between encodings for the infraction values of an expert and the current policy.
\begin{equation} \label{eqn3}
	L_{It} = \|S_t - \hat{S_t}\|^{2} + KL(q_I(z_I)||p_I(z_I)),
	\end{equation}
with \textbf{$p_I(z_I)$} following the same normal distribution as \textbf{$p_D(z_D)$}.

With these reward functions, the dynamic reward $\textbf{R}_D$, the safety reward $\textbf{R}_S$ and the task reward $\textbf{R}_{T}$ (that is a function of the distance of bucket from the goal location), we aim to learn a policy $\pi$ for excavator automation in a complex workspace. Given the actuator state configuration $\textbf{s}$ (described in Section 4) and the action space $\textbf{a}$, policy $\pi$ parameterized as $\theta$ is used to sample the action,
\begin{equation} \label{eqn4}
	a = \pi_{\theta}(s)
	\end{equation}
such that it maximizes the expected discounted reward $\textbf{J}$,
\begin{equation} \label{eqn5}
	J(\pi) = E_{\pi_{\theta}}[\sum_{t=0}^{t=m-1}\gamma^{t} r(s_t,a_t)]
	\end{equation}

Here $\gamma \in (0,1]$ is the discount factor, $t$ is the current time step of the episode, $r$ is the total reward considering the different reward functions and $m$ is the total time steps in a given episode. 
	
Both latent space $\textbf{z}$ and policy $\pi$ are modelled using a neural network. $\textbf{q}_D\textbf{(z)}$ uses a Long Short Term Memory (LSTM) \citep{hochreiter1997long} while $\textbf{q}_I\textbf{(z)}$ and policy $\pi$ uses a simple 2-layered deep multi-layer perceptron. The methodology and the results are further elaborated in Sections 4 and 5 respectively. Considering the limited work in this direction we compare the use of our proposed reward function for learning a policy by considering a policy trained using only a task-based reward function as a baseline.

\begin{figure}[!ht]
  \centering
  \includegraphics[scale=0.35]{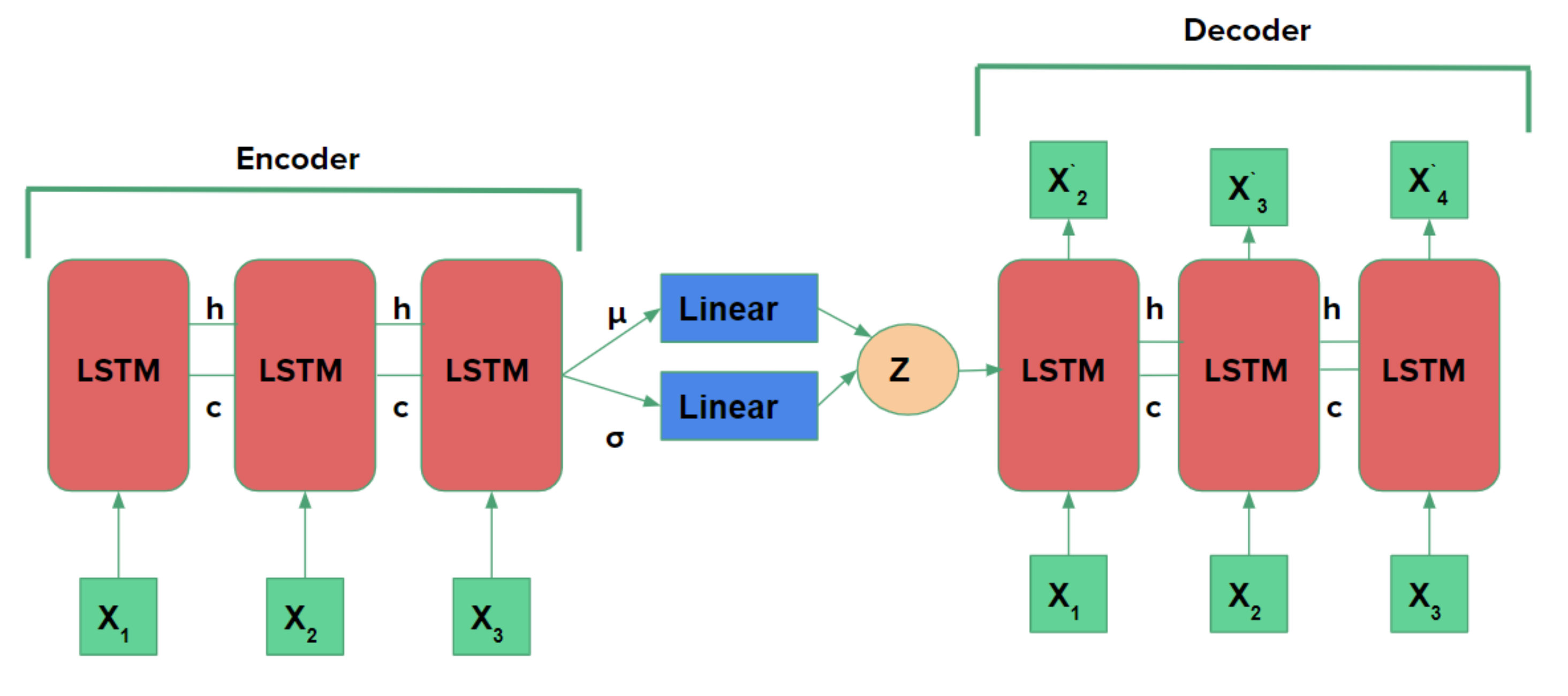}
  \caption{\textbf{Model for learning the dynamic distribution} The latent vector \textbf{z} encoding the expert's dynamic distribution is learned using LSTM-based encoder-decoder by optimizing for the task of future state prediction. During training, the decoder uses the original last state (teacher-forcing strategy) and predicts the future state using LSTM by initializing its hidden state and the cell state with the dynamics predicted by the encoder.}
  \label{figurelabel3}
\end{figure}
\section{Method}
For tasks like navigation and manipulation, current reinforcement learning algorithms mostly use goal-based rewards \citep{nair2018visual},\citep{nasiriany2019planning}, maximising the cumulative rewards over a given time. This strategy does not take into account the machine dynamics or the external environmental constraints limiting its use for real-life applications. In this work, we address this problem by proposing additional reward signals along with the original goal-based reward. These rewards are learned using the distribution of expert demonstrations. We limit this distribution to learning the representation of certain features only, which needs to be taken into consideration by the excavator while solving the main goal. Our final aim is to use these rewards to learn an RL policy to solve the challenging task of excavator maneuvering in a constrained environment. 
Fig. \ref{figurelabel} summarizes our automatic reward prediction framework for the given task. \newline \\
\textbf{Learning Distributions of Expert's Demonstrations}. Our model learns two independent distributions: a temporal representation of different excavator features and a non-temporal distribution of environmental states (Table \ref{tab:1}) given demonstrations of multiple experts. The given temporal features are directly impacted by the user's actions, efficiently modelling the dynamic distribution, while the infraction values at every time step take into account the excavator's interaction with the environment, acting as an ideal source for modelling the safety distribution. Other observations like different actuator trajectories and their velocity are not included in the learned distribution forcing the reinforcement learning policy to discover its own novel strategies based on the required goal. 

To learn the dynamic distribution, for a given time step $\textbf{t}$ we sample the last 32 readings for each feature in the form of 32 x 3 time-series data. This sequence of temporal data is encoded into a single 8-d feature vector using Long Short Term Memory Network (LSTM). This representation is then used to predict the mean ($\mu$) and variance ($\sigma$) for a given distribution using two independent Multi-Layer Perceptrons (MLPs). Further, to encode the dynamics of the given temporal features in the predicted distribution $\textbf{q}_D\textbf{(z)}$, the objective formulated requires the decoder to predict the feature values for the next time steps given the sampled vector from the distribution and the feature values of the last time step as input. This decoder consists of LSTM which takes as input the original feature values for the last time step and predicts the future values. To integrate user dynamics we initialise the hidden and cell state of LSTM with the sampled vector from the predicted distribution. The overall architecture as shown in Fig. \ref{figurelabel3} is trained using the reconstruction loss of the predicted future states. Along with this, an additional regulariser in the form of KL Divergence loss is added such that the predicted distribution is Normal Distributed with a mean of 0 and variance 1.

\begin{figure}
    \centering
    \subfloat[Network Architecture]{\includegraphics[width=6cm,height=6cm,keepaspectratio]{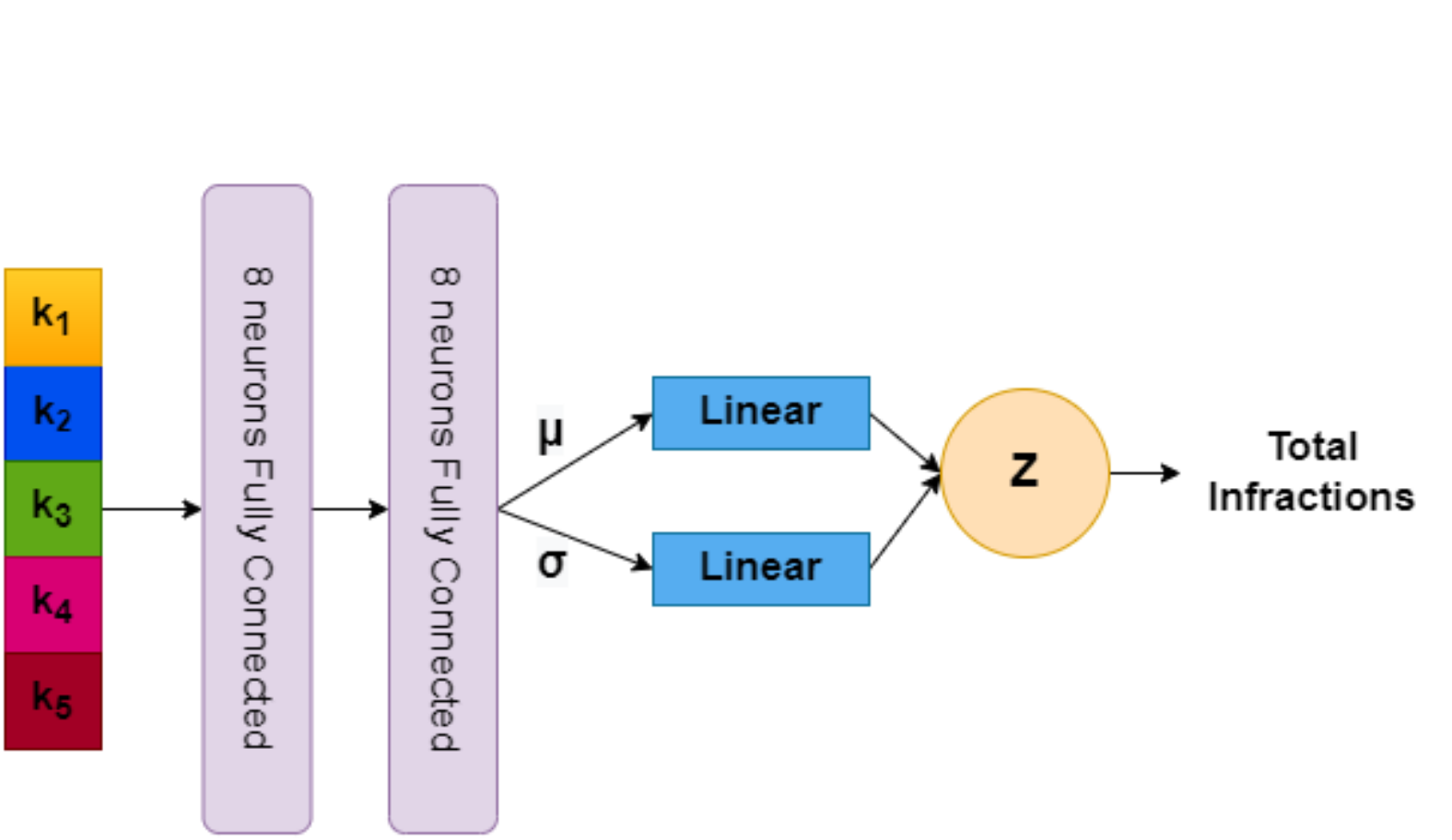}\label{fig:safnet}}
    \subfloat[Safety Distribution]{\includegraphics[width=5cm,height=5cm,keepaspectratio]{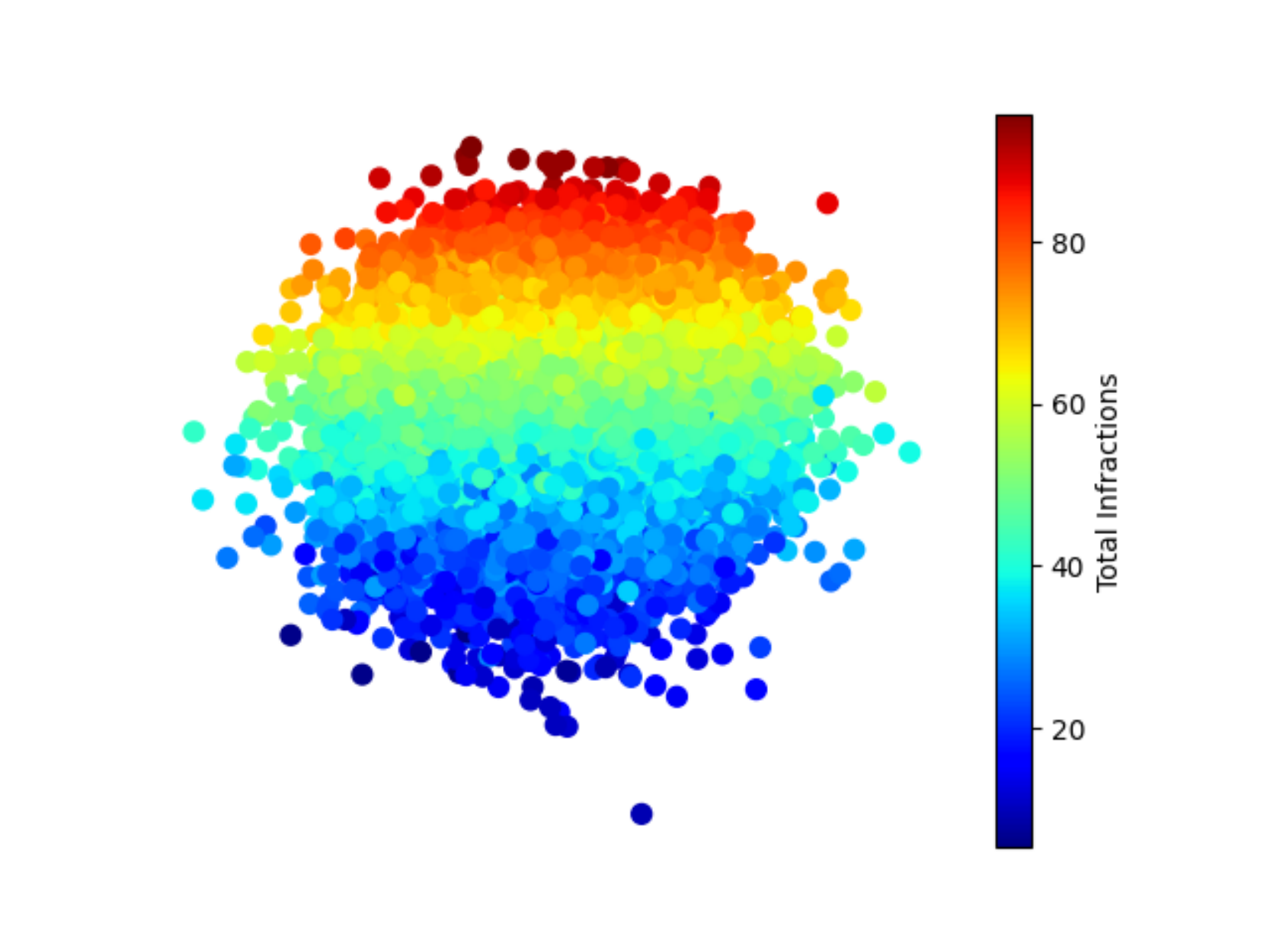}\label{fig:safdist}}
    \caption{\textbf{Model for learning the safety distribution} By learning a distribution using the objective of predicting the total infractions the encodings efficiently differentiate between two infractions hence acting as an optimal reward function to integrate safety into the given policy.}
    \label{fig:Infraction-distribution}
\end{figure}

The safety distribution $\textbf{q}_I\textbf{(z)}$ is modelled following a similar strategy where a 5-d feature vector of infraction values is encoded into a 2-d encoding representing the distribution using a 2-layered (8 neurons) fully connected network as shown in Fig. \ref{fig:safnet}. This is optimized using the objective of predicting the sum of the total infractions given the input along with the KL divergence loss (no reconstruction loss). As shown in Fig \ref{fig:safdist} this approach effectively differentiates between infractions with widely different values.

These distributions of experts' actions are then used to model the reward function. Given a new user, for a certain action at a given time step $t$ the learned model (weights frozen) takes as input the values of the features as specified in Table \ref{tab:1} and maps its distribution. The reward for the action taken is then evaluated using the KL Distance between the predicted and the expert's distribution. This is further summed  with the original task-based reward for training an RL policy.

\begin{table}
\begin{center}
\begin{tabular}{c|c|c}
\multicolumn{1}{c|}{\textbf{Dynamic}}&\multicolumn{2}{c}{\textbf{Safety}}
 \\ \cline{2-3}
 \hline
 Average Engine Torque & Poles Touched & Environment Collision\\
 \hline
 Average Engine Power &  Poles Fell & Balls knocked\\ 
 \hline
 \multicolumn{1}{c|}{Average Fuel Consumption} & \multicolumn{2}{c}{Equipment Collision}\\
 \hline
\end{tabular}
\end{center}
\caption{\label{tab:1}Inputs for Modelling the Dynamic and Safety Distribution.}
\end{table}

\textbf{Policy Learning.} The final aim is to learn an RL policy for solving the task of maneuvering an excavator in a complex environment. This task is challenging to model using classical controllers as it requires precise coordination of multiple actuators. Also, it is difficult to adapt these controllers \citep{Jud2019AutonomousFT},\citep{Koivo1996ModelingAC} across different excavators considering the wide variations. Reinforcement learning is an ideal solution as it forces each actuator to coordinate its actions to discover optimal policies \citep{liu2021deep}. Moreover, these policies are robust \citep{almasi2020robust} to variations,  providing a scalable solution for real-life adoption. Considering its stability and robustness to different hyperparameters, we use a model-free Proximal Policy Optimization (PPO) \citep{schulman2017proximal} an on-policy algorithm. This policy is modelled using a 2-layered MLP where the input space includes a 19-d vector as shown in Table \ref{tab:3}. The output is a continuous 4-d vector within a range of [-1,1] which represents the setpoints for each actuator, such that values less than 0 represent the movement of the actuators in the negative direction, while values greater than 0 move the actuator in the positive direction. For the swing actuator, it is equivalent to rotation in the left or right direction. To optimise this policy for the excavator to complete the specified goal we use the proposed reward functions that is the dynamic reward ($R_D$) and the safety reward ($R_S$) along with the distance-based reward ($R_G$), where the distance is calculated between the bucket's spatial coordinates and the specified goal as shown in Eqn \ref{eqn6}. For both the dynamic reward and the safety reward, we use the pre-trained model to predict the distribution $p_{AD}$ and $p_{AI}$ for the given time step.

\begin{equation} \label{eqn6}
	R = R_G + R_D + R_S 
	\end{equation}
\quad $R_G = 1 - ||G - B_L||$;\quad
$R_D = 1 - KL(p_{AD}(z_D)||p_{D}(z_D))$; \quad
$R_S = 1 - KL(p_{AI}(z_I)||p_{I}(z_I))$
\begin{table}[htp]
\begin{center}
\begin{tabular}{c|c|c|c|c|}
\multicolumn{1}{c|}{Inputs}&\multicolumn{4}{c|}{Actuators}\\ \cline{2-5}
&Swing&Boom&Stick&Bucket\\ \hline
Linear Position&\CheckmarkBold&\CheckmarkBold&\CheckmarkBold&\CheckmarkBold\\ \hline
Angular Velocity&\CheckmarkBold&\CheckmarkBold&\CheckmarkBold&\CheckmarkBold\\ \hline
Linear Velocity&\textbf{--}&\CheckmarkBold&\CheckmarkBold&\CheckmarkBold\\ \hline
\end{tabular}
\end{center}
\caption{\label{tab:3}The observation space of the reinforcement learning policy.}
\end{table}

\section{Experiments}
Through our experiments, we aim to verify: 
\begin{enumerate}[leftmargin=0.5cm]
    \item The effectiveness of using expert distribution for modelling the reward function.
    \item The impact of learning the RL policy for the task of excavator maneuvering using these additional rewards formulated as dynamics and infraction distribution in comparison to task-based reward.
\end{enumerate}

\textbf{Simulator Setup and Dataset Collection.} For collecting both the user data as well as for training the reinforcement learning policy Vortex simulator\footnote{https://www.cm-labs.com/vortex-studio/} is used. This simulator provides a range of customisable construction environments for different crane automation tasks. In our case, we use the scenario shown in Fig. \ref{figurelabel2}, where the task is to maneuver the crane to reach a pre-specified goal location. We fixed the goal and start location as shown in Table \ref{tab:4} in a way to provide a complex challenging task in the presence of all the infractions for the crane to complete for proper analysis. 

For our dataset, we collect trajectories for a total of 40 users where each is ranked based on pre-defined heuristics. This heuristic is formulated such that a user is automatically penalised in case of any infractions. The final score ranges from 0 (maximum) to -100 (minimum) for each session. We consider a user to be an expert if the total score is greater than -25, while others are considered a novice. In the current dataset, 7 users have a score greater than -25 and are further used to learn the distributions. \newline
\\
\textbf{Implementation.} The model for learning the distributions is trained using the trajectories of 5 experts and the remaining trajectories are used for evaluating the model. Both the LSTM-based encoder-decoder setup and the 2 layered fully connected network is trained for a total of 1000 epochs with a batch size of 8 using an Adam Optimizer with a learning rate of 1e-4. Similarly, for policy training using an actor-critic architecture, the networks are trained for 500 episodes with a learning rate of 3e-4 for the actor and 1e-3 for the critic. For all the layers other than the output we used the ELU activation function
\begin{table}[!h]
\begin{center}
\begin{tabular}{|c|c|c|c|} 
\hline
 Complexity & Start & Goal & Distance (in m) \\
 \hline
  1 &(-9.8,-144.1,2.4)&(-2, -150, 0.5) & 10.2\\
 \hline
\end{tabular}
\end{center}
\caption{\label{tab:4}Description of Task to learn.}
\end{table}
\cite{clevert2015fast} as it showed better performance and faster learning. For a given episode the RL policy  runs for a maximum of 280 steps and the episode terminates on reaching the goal. The networks are trained on an Nvidia GTX 1050 Ti with Intel Core i5-8300H CPU @ 2.30GHz × 8, which takes around 2 to 3 hours for learning the distributions and 4 to 5 hours for training the policy. \\
\newline
\textbf{Results.} We evaluate the impact of using our proposed reward model on the final policy by comparing different dynamic features and the infraction values as shown in Table \ref{tab:5}. Each policy is trained with and without the different reward functions along with the original task reward. The excavator in the absence of any constraints performs the worst when compared with the expert (see values of Task Reward in Table \ref{tab:5}). With the addition of constraints in the form of the proposed reward function the performance drastically improves to follow the expert's dynamics and infraction values. It is interesting to observe the performance of the policy when both the dynamic and safety rewards are included, both showing almost similar performance much better than the policy trained with only task reward. The safety reward optimises the policy, for the infractions (Table \ref{tab:5}), reducing the infractions to zero (a safe policy) while using it as the only reward does not take into account the dynamics. A similar trend is observed when only dynamic reward is used.

\begin{figure}
    \centering
    \subfloat[Task Reward - Final Distance of the bucket from Goal (less is better) ]{\includegraphics[width=5cm,height=5cm,keepaspectratio]{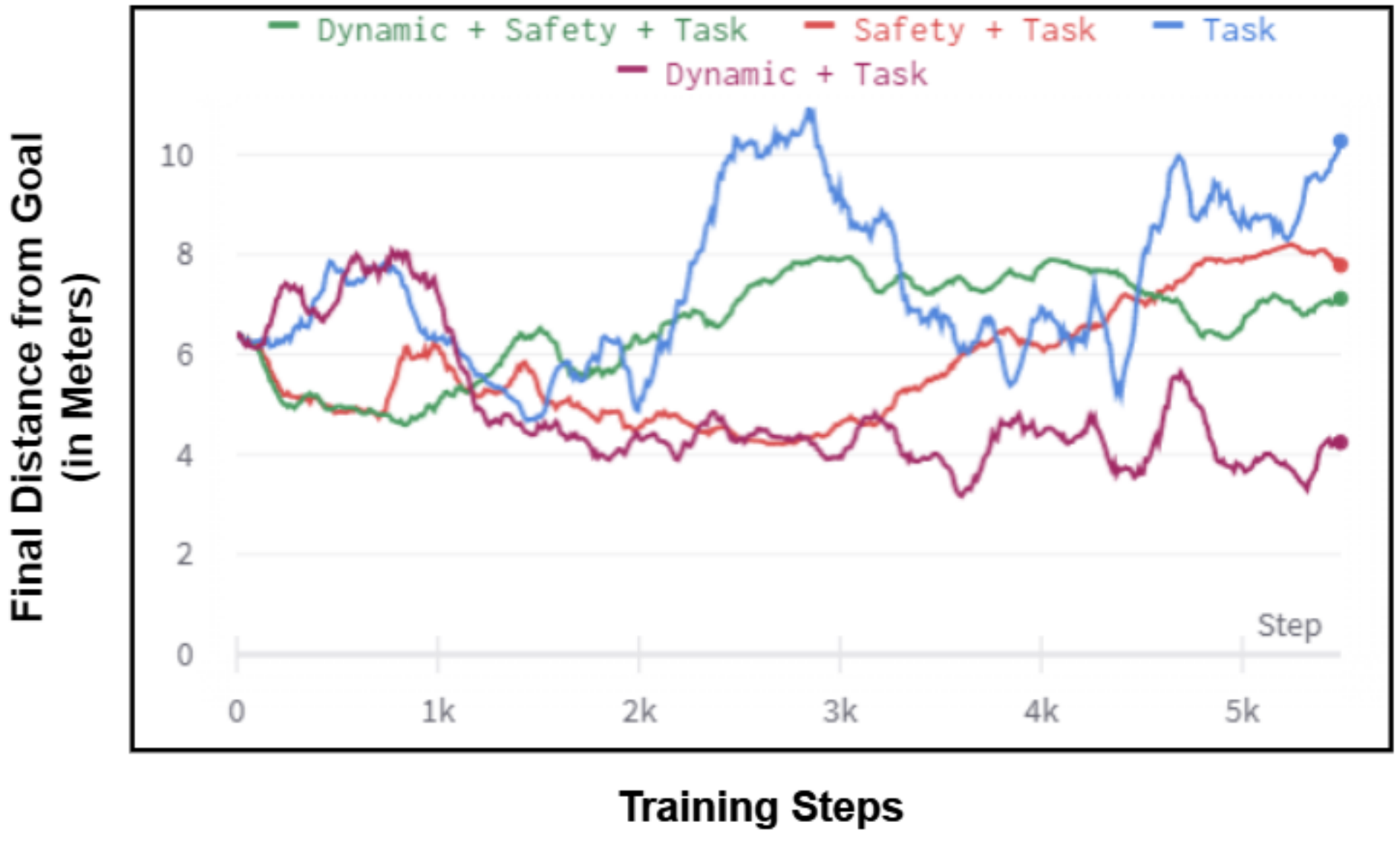}\label{fig:reward1}}
    \subfloat[Dynamic Reward]{\includegraphics[width=4.5cm,height=4.5cm,keepaspectratio]{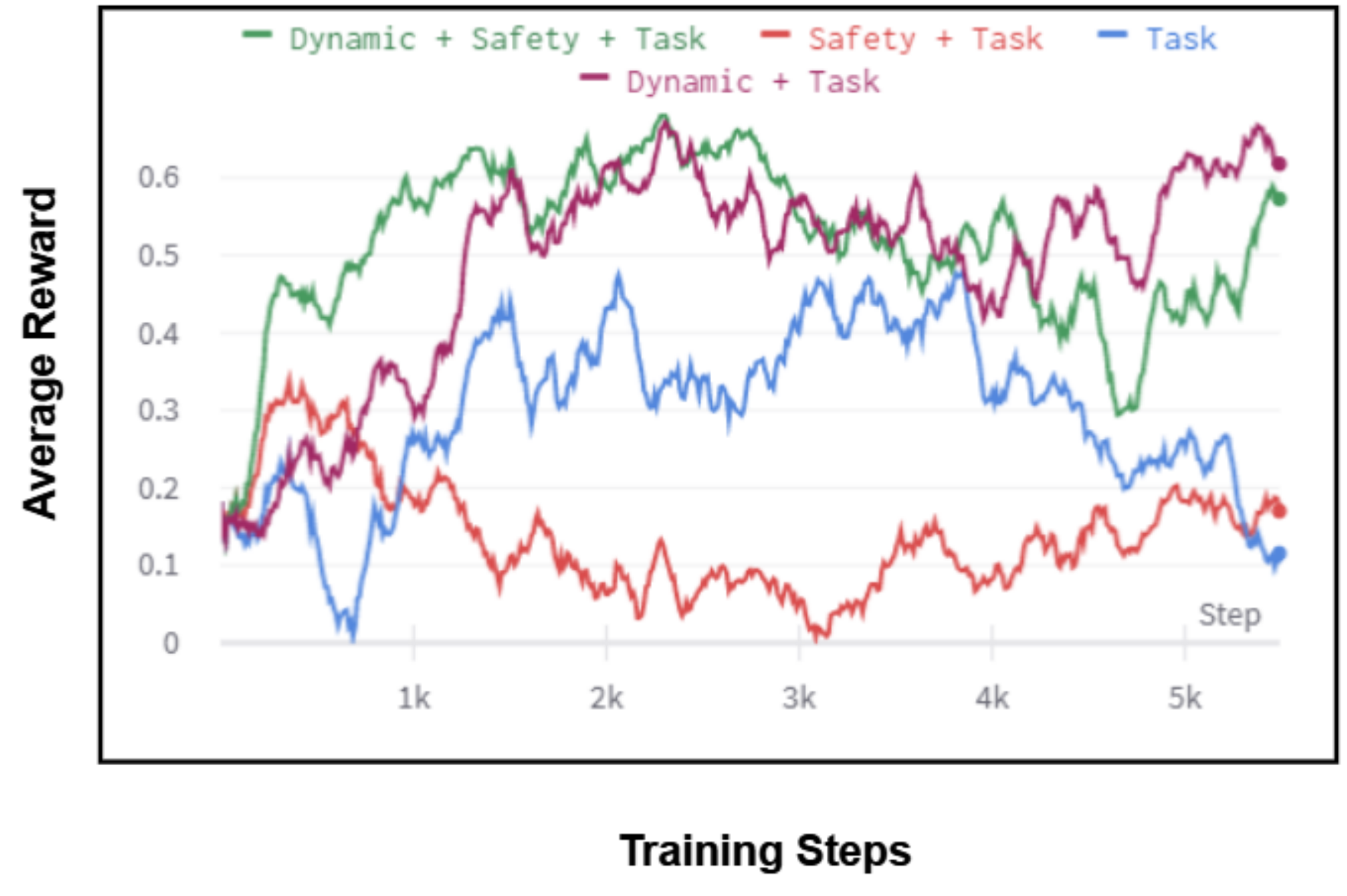}\label{fig:reward2}}
    \subfloat[Safety Reward]{\includegraphics[width=5cm,height=5cm,keepaspectratio]{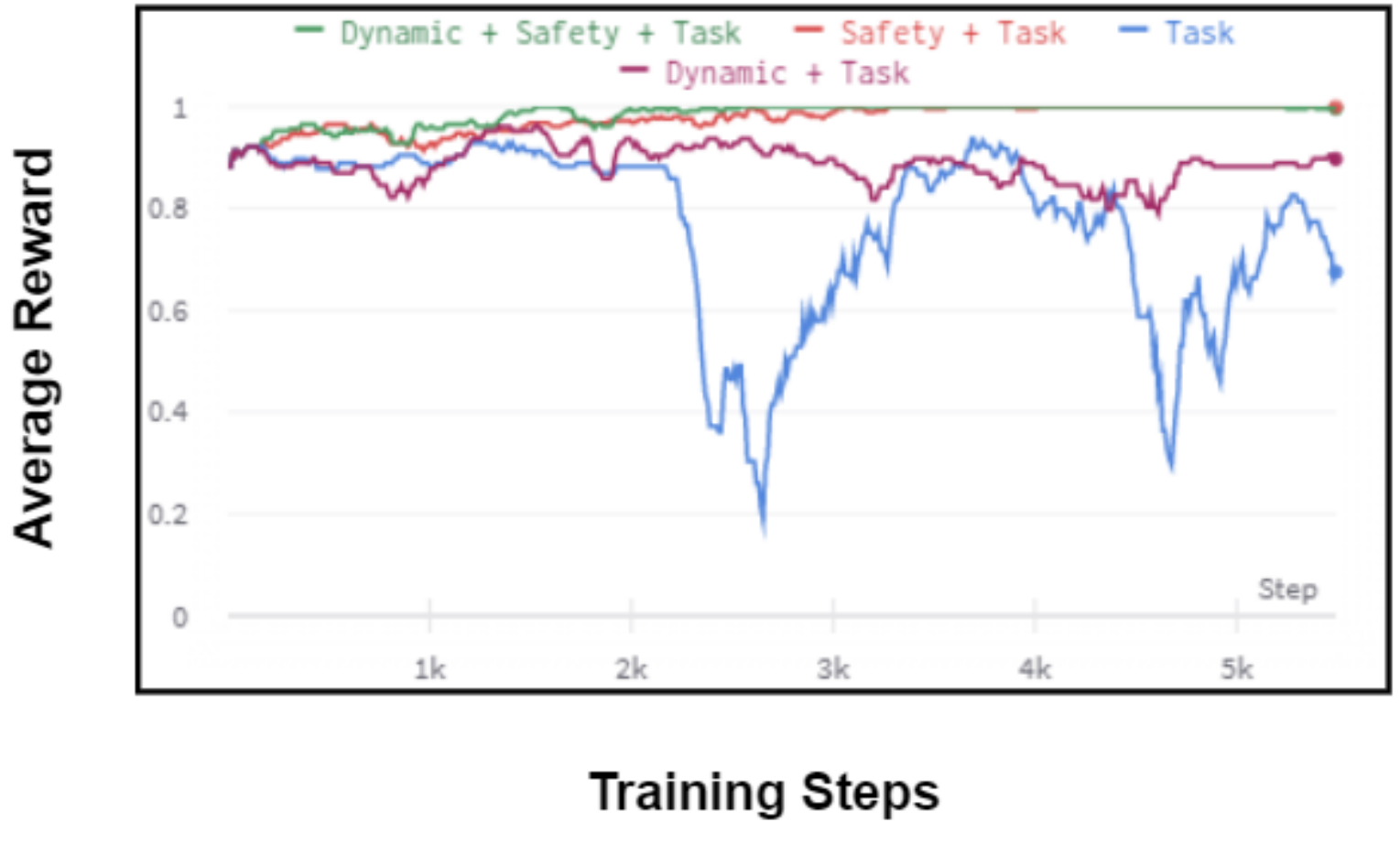}\label{fig:reward3}}
    \caption{Impact of training the policy using different reward functions.}
    \label{fig:graphs_ablation}
\end{figure}

In Fig.\ref{fig:graphs_ablation}, the optimisation of policy for different rewards is compared. For the task reward, (Fig. \ref{fig:reward1}) which is the original goal of reaching the final specified position, the best performance is achieved with the addition of a dynamic reward, better than the task reward alone. This is because by adding a dynamic reward with the task reward, the excavator is constrained to the dynamic features of an expert, preventing it to explore unwanted actions and leading to faster convergence. Policy's dynamics, (Fig. \ref{fig:reward2}) experience a significant drop when trained without the inclusion of dynamic reward. A similar trend is observed for policy trained without any safety reward (Fig. \ref{fig:reward3}). Comparing each constraint, only using task rewards for learning, shows the worst performance while for the policies integrated with additional constraints, the performance closely follows that of an expert.
\begin{table}[!h]
    \scriptsize
    \begin{center}
    \setlength{\tabcolsep}{0.033\linewidth}
    \begin{tabular}{cccccc}
    \toprule
       
          & \textbf{Expert} & Task & \textbf{Dynamic} & Safety & \textbf{Dynamic+Safety}\\\cline{2-6}\\
         Avg. Torque (in \%) & \textbf{24-34}& 51.73 & \textbf{45.61}& 51.28& \textbf{46.44}\\
         Avg. Power (in \%) & \textbf{10-20}& 26.77 & \textbf{22.63}& 26.84& \textbf{23.72}\\
         Avg. Fuel Consumption (in \%) & \textbf{3-5}& 5.77 & \textbf{5.01}& 5.71& \textbf{5.12}\\
         Total Infractions & \textbf{0}& 35 & 18& \textbf{0}& \textbf{1}\\
         \midrule
    \end{tabular}
    \end{center}
    \caption{\label{tab:5}The Dynamic and the Infraction values of the final policy for different reward functions.} 
\end{table}
\section{Conclusion}
\textbf{Summary.} In this work, we propose a novel approach for modelling a feedback mechanism in the form of multiple independent distributions  using demonstrations of expert operators to teach new students the task of excavator maneuvering. These distributions that are the dynamic and the safety distribution are learned using the temporal states of the excavator and infraction values of the expert's trajectory. Once learned, the distributions are used as feedback for trainees by taking into account the KL divergence with the current action. This automatic score predictive model is further validated by using it as a reward mechanism for the RL policy  to learn the task in a simulated environment.  We show that the policy learned using this additional supervision completes the goal (using the task reward) while also optimising for the additional constraints. The final policy successfully maneuvers the excavator while closely emulating the dynamics and the infraction values of an expert operator.

\textbf{Limitations and Future Work.} For future work, we plan to explore additional constraints by learning a richer representation. The current strategy is computationally inefficient as it requires learning independent distributions for each additional constraint, instead, a single complex model may lead to significant improvements. Along with this, we plan to extend our work for a diverse range of challenging real-life applications for different construction tasks like loading-unloading, soil excavation and levelling.

\section{Acknowledgements}
We would like to thank Mitacs and CMLabs for funding the project. Further, we are grateful to Hsiu-Chin Lin for the valuable feedback during the initial phase. We are also thankful to the Digital Research Alliance of Canada for the computing resources and CIFAR for research funding.

\bibliographystyle{plainnat}
\bibliography{example}

\end{document}